\documentclass[letterpaper, 10 pt, conference]{styles/ieeeconf}
\IEEEoverridecommandlockouts                             
\usepackage{url}
\usepackage{color}
\usepackage{wrapfig}
\newif\iffigs
\figstrue % comment out to skip compiling figures

\usepackage[fleqn]{amsmath}
\usepackage{float}
\usepackage{setspace}
\usepackage{graphicx}
\usepackage{mathrsfs}
\usepackage{amssymb}
\usepackage{nicefrac}
\usepackage{algorithm}
\usepackage[noend]{algpseudocode}

\usepackage{flushend}
\usepackage{multicol}
\newcommand\nocell[1]{\multicolumn{#1}{c|}{}}
\usepackage{hyperref}
\hypersetup{bookmarks=true}

\makeatletter
\newcommand\fs@spaceruled{\def\@fs@cfont{\bfseries}\let\@fs@capt\floatc@ruled
  \def\@fs@pre{\vspace{0.4\baselineskip}\hrule height.8pt depth0pt \kern2pt}%
  \def\@fs@post{\vspace{-0.4\baselineskip}\kern2pt\hrule\relax\vspace{-12pt}}%
  \def\@fs@mid{\kern2pt\hrule\kern2pt}%
  \let\@fs@iftopcapt\iftrue}
\makeatother

\title{\LARGE \bf GRiD: GPU-Accelerated Rigid Body Dynamics\\with Analytical Gradients}

\author{Brian Plancher$^{1}$, Sabrina M. Neuman$^{1}$, Radhika Ghosal$^{1}$, Scott Kuindersma$^{1,2}$, Vijay Janapa Reddi$^{1}$% <-this % stops a space
\thanks{This material is based upon work supported by the National Science Foundation 
(under Grant DGE1745303 and Grant 2030859). % to the Computing Research Association for the CIFellows Project). 
Any opinions, findings, conclusions, or recommendations expressed in this material are 
those of the authors and may not %do not necessarily 
reflect those of the funding organizations.}% <-this % stops a space
\thanks{$^{1}$Brian Plancher, Sabrina M. Neuman, Radhika Ghosal, Scott Kuindersma, and Vijay Janapa Reddi are with the John A. Paulson School of Engineering and Applied Sciences, Harvard University, Cambridge, MA. {\tt\footnotesize \{brian\_plancher@g, rghosal@g, sneuman@seas, scottk@seas, vj@eecs\}.harvard.edu}}%
\thanks{$^{2}$Scott Kuindersma is also with Boston Dynamics, Waltham, MA.}%
}

\begin{document}
\maketitle
\thispagestyle{empty}
\pagestyle{empty}

%%%%%%%%%%%%%%%%%%%%%%%%%%%%%%%%%%%%%%%%%%%%%%%%%%%%%%%%%%%%%%%%%%%%%%%%%%%%%%%%

\begin{abstract}
    We introduce GRiD: a GPU-accelerated library for computing rigid body dynamics with analytical gradients. 
GRiD was designed to accelerate the nonlinear trajectory optimization subproblem used in state-of-the-art robotic planning, control, and machine learning, 
which requires
% Each iteration of nonlinear trajectory optimization requires 
tens to hundreds of naturally parallel computations of rigid body dynamics and their gradients
at each iteration.
GRiD leverages URDF parsing and code generation to deliver optimized dynamics kernels that not only expose GPU-friendly computational patterns, but also take advantage of both fine-grained parallelism within each computation and coarse-grained parallelism between computations. 
Through this approach, when performing multiple computations of rigid body dynamics algorithms, GRiD provides as much as a 7.2x speedup over a state-of-the-art, multi-threaded CPU implementation, and maintains as much as a 2.5x speedup when accounting for I/O overhead.
% We release GRiD as an open-source library, so that it can be leveraged by the robotics community to easily and efficiently accelerate rigid body dynamics on the GPU.
We release GRiD as an open-source library for use by the wider robotics community.
\end{abstract}

\section{Introduction} \label{sec:intro}
Efficient implementations of rigid body dynamics and their gradients have become key computational kernels for robotics applications. Originally required mostly for the nonlinear trajectory optimization sub-problems of model-based planning and control systems for high degrees-of-freedom robots~\cite{Farshidian17,Giftthaler17,Plancher19a,Kleff20}, these computational kernels are also growing in importance for machine learning (ML) techniques~\cite{Carius20,Omer20,Hoeller20,Gros20,Morgan21}.  

Despite being highly accurate and optimized, existing implementations of spatial-algebra-based approaches to rigid body dynamics~\cite{Featherstone08} do not take advantage of opportunities for parallelism present in the algorithm, limiting their performance~\cite{Frigerio16,Carpentier18,Carpentier19,Neuman19}.
This is critical because there is natural parallelism in many bottleneck computations involving rigid body dynamics in robotics~\cite{Williams17,Plancher18,Morgan21}. 
For example, the gradient of forward dynamics accounts for $30$\% to $90$\% of %the total computational time of 
typical nonlinear model-predictive control (MPC) implementations~\cite{Koenemann15,Neunert16b,Carpentier18,Plancher18}, and is naturally parallel across the discrete points in the trajectory.

While there is a growing need for parallel computation, the performance of multi-core CPUs has been limited by thermal dissipation, enforcing a utilization wall that restricts the performance a single chip can deliver~\cite{Esmaeilzadeh11,Venkatesh10}.
This has motivated increased use of GPUs, which can provide opportunities for higher performance by supporting larger-scale parallelism within a single chip.

In this work, we introduce \emph{GRiD}, a GPU-accelerated library for spatial-algebra-based rigid body dynamics and their analytical gradients. GRiD is designed to accelerate the nonlinear trajectory optimization subproblem used in state-of-the-art robotic planning, control, and machine learning algorithms. GRiD is optimized to use blocks of GPU threads to compute the tens to hundreds of naturally parallel computations of rigid body dynamics and their gradients found in these algorithms, and implements the more accurate spatial-algebra-based formulation of rigid body dynamics used in state-of-the-art trajectory optimization~\cite{Erez15,Sleiman2021,Drnach2021,Freeman21}. GRiD builds on recent work which designed a manually-optimized GPU implementation of rigid body dynamics gradients for a seven-link serial chain manipulator~\cite{Plancher21}, providing further optimizations and generalizations to support multiple dynamics algorithms, \texttt{URDF} parsing of most common robot models, and optimized code generation.

GRiD not only unlocks the ability for nonlinear trajectory optimization to run entirely on the GPU, but when performing multiple computations of rigid body dynamics and their gradients, it also provides as much as a 7.2x speedup over a state-of-the-art, multi-threaded CPU implementation running on a high-performance workstation. GRiD also enables the use of a GPU as a rigid body physics accelerator for algorithms that are computed on a host CPU, maintaining as much as a 2.5x speedup when accounting for the I/O communication overhead between the CPU and GPU.

We release GRiD as an open-source library to enable robotics researchers to better explore and leverage the performance gains from large-scale parallelism on GPU platforms. Our library can be found at {\small \href{https://github.com/robot-acceleration/grid}{\color{blue} \texttt{https://github.com/robot-acceleration/grid}}.}

\section{Related Work} \label{sec:related}
GRiD is designed to provide general-purpose, spatial-algebra-based dynamics with analytical gradients, and to accelerate them through large-scale parallelism on the GPU.

While there are many existing state-of-the-art spatial-algebra-based rigid body dynamics libraries~\cite{Todorov12,Frigerio16,Carpentier19,Koolen19,Werling2021}, these libraries are not optimized for GPUs~\cite{Plancher18}. 
The exception is the recently released NVIDIA Isaac Sim~\cite{makoviychuk21} which supports spatial-algebra-based forward simulation, but not gradients, on the GPU.

As such, prior work using spatial-algebra-based approaches for planning and control on GPUs were either limited to cars, drones, and other lower degrees-of-freedom systems~\cite{Williams17}, or relied on manually-optimized implementations of rigid body dynamics and their gradients for a specific robot model~\cite{Plancher18,Plancher21}. Most machine learning approaches that leverage spatial-algebra-based rigid body dynamics rely on these aforementioned libraries~\cite{Morgan21,makoviychuk21}.

Using automatic differentiation, differentiable physics engines can also support gradient computations and have shown promise for real-time nonlinear MPC use on CPUs~\cite{Neunert16a} and for accelerating machine learning, computer graphics, and soft robotics applications~\cite{Bender17,DeAvila18,Degrave19,Hu19,Austin20,Hu20,Freeman21,Heiden2021a} on CPUs and GPUs. However, existing GPU-based differentiable physics engines are optimized for simulating thousands of interacting bodies through contact using maximal coordinate, particle, and mesh-based approaches, which are less accurate when used for rigid body robotics applications over longer time step durations~\cite{Erez15,Freeman21}.

GPUs have also historically been used to accelerate gradient computations through numerical differentiation~\cite{Micikevicius09,Michea10}. However, these methods have been shown to have less favorable numerical properties when used for robotic planning, control, and machine learning.

\section{Background} \label{sec:background}
\subsection{Computing Hardware: CPUs vs. GPUs} \label{sec:gpu_background}

Compared to a multi-core CPU, a GPU has a much larger set of simpler processors, optimized specifically for parallel computations with identical instructions operating over data accessed in regular patterns. %(SIMD parallelism).
Each GPU processor has many more arithmetic logic units (ALUs), but reduced control logic and a smaller cache memory (see Figure~\ref{fig:cpu_gpu}). 
%For maximal performance, the GPU requires groups of threads within each thread block to compute the same operation on memory accessed via regular patterns.
GPUs are therefore best at computing highly regular and separable computations over large working sets of data (e.g., large matrix-matrix multiplication) where much of the cache, referred to as \textit{shared memory}, can be manually manged by the programmer. %It is also worth noting that GPUs typically run at about half the clock rate of CPUs, which further hinders their performance on purely sequential code. 
We also note that data must be transferred between the CPU and GPU  incurring I/O overhead.
% When leveraging a GPU as an accelerator, for each independent computation, data must be transferred from the CPU's memory to the GPU and then back again after computations are completed. This I/O communication overhead can be amortized by performing large amounts of arithmetic operations on the GPU per each round trip memory transfer. From a design perspective, GPUs are best suited for applications which require high throughput of the compute workload, have high computational intensity, and exhibit high degrees of natural parallelism.

\iffigs
\begin{figure}[ht]
   \centering
%   \vspace{5pt}
   \includegraphics[width=0.9\columnwidth]{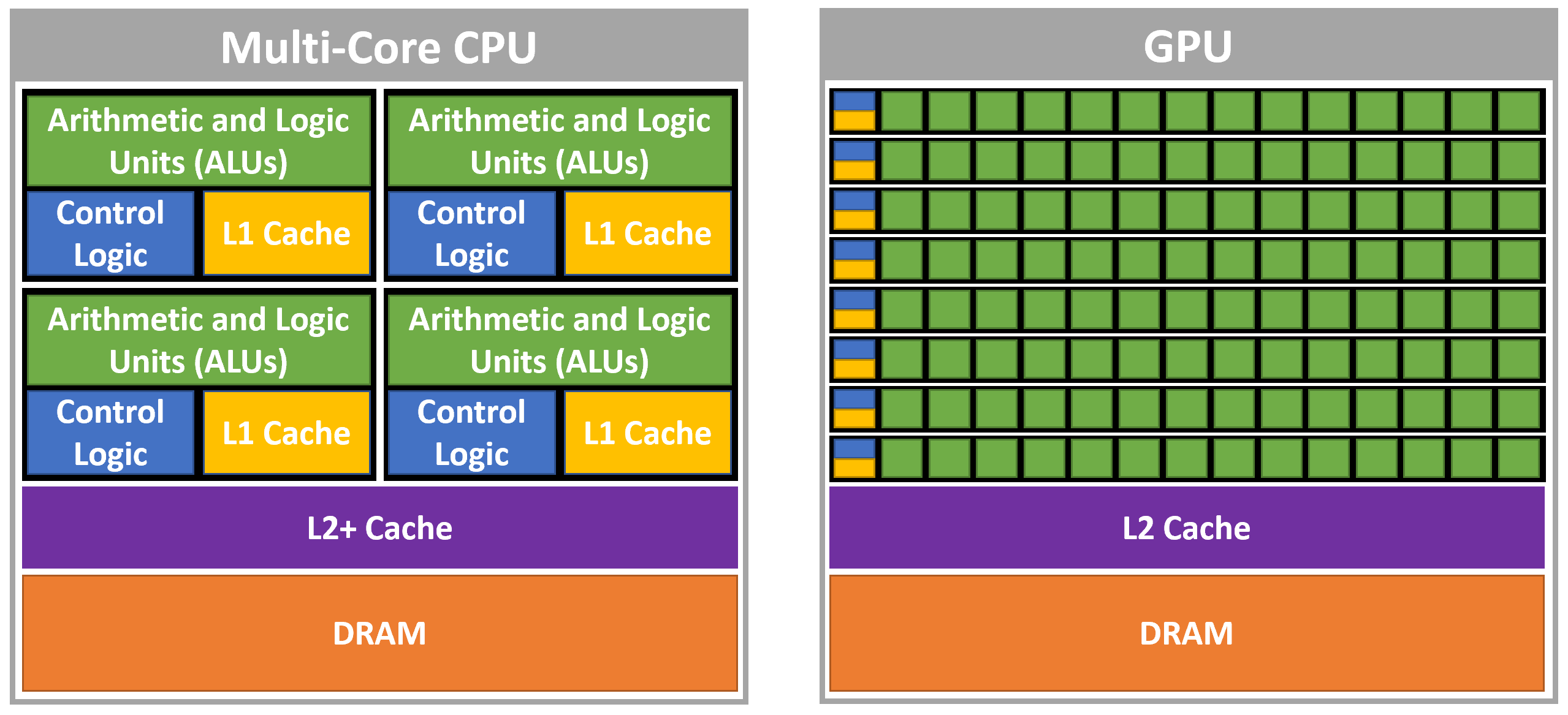}
%   \vspace{-15pt}
   \caption{High level architecture of a multi-core CPU and GPU processor.} % Inspired by~\cite{NVIDIA18}.}
  \vspace{-2pt}
   \label{fig:cpu_gpu}
\end{figure}
\fi

Our work uses NVIDIA's CUDA~\cite{NVIDIA22} extensions to C++ which uses parallel blocks of threads to compute functions on the GPU. Each block's threads access a shared cache and are guaranteed to run on the same processor, but the ordering of the blocks is not guaranteed. For more information on CUDA and its programming model we suggest reading the NVIDIA CUDA programming guide~\cite{NVIDIA22}.
% Our work uses NVIDIA's CUDA~\cite{NVIDIA18} extensions to C++. CUDA is built around \textit{host} (CPU) and \textit{device} (GPU) memory and code. Special functions called \textit{kernels} are launched from the host and then call device functions using parallel blocks of threads on the GPU. Each block's threads access a shared cache and are guaranteed to run on the same processor, but the ordering of the blocks is not guaranteed. For more information on CUDA and its programming model we suggest reading the NVIDIA CUDA programming guide~\cite{NVIDIA18}.

%%%%%%%%%%%%%%%%%%%%%%%%%%%%%%%%%%%%%%%%%%%%%%%%%%%%%%%%%%%%%%%%%%%%%%%%%%%%
%%%%%%%%%%%%%%%%%%%%%%%%%%%%%%%%%%%%%%%%%%%%%%%%%%%%%%%%%%%%%%%%%%%%%%%%%%%%
%%%%%%%%%%%%%%%%%%%%%%%%%%%%%%%%%%%%%%%%%%%%%%%%%%%%%%%%%%%%%%%%%%%%%%%%%%%%
%%%%%%%%%%%%%%%%%%%%%%%%%%%%%%%%%%%%%%%%%%%%%%%%%%%%%%%%%%%%%%%%%%%%%%%%%%%%

\subsection{Rigid Body Dynamics} \label{sec:rbd_background}

State-of-the-art spatial-algebra-based rigid body dynamics algorithms~\cite{Featherstone08} operate in minimal coordinates and compute functions of the joint position $q \in \mathbb{R}^{n}$, velocity $\dot{q} \in \mathbb{R}^{n}$, acceleration $\ddot{q} \in \mathbb{R}^{n}$, and input torque $\tau \in \mathbb{R}^{m}$ that satisfy:
\begin{equation} \label{eq:manipulator}
    M(q)\ddot{q} + C(q,\dot{q})\dot{q} + G(q) = B(q)\tau + J(q)^T F
\end{equation}
where $M(q) \in \mathbb{R}^{n \times n}$ is the mass matrix, $C(q,\dot{q}) \in \mathbb{R}^{n \times n}$ is a Coriolis matrix, $G(q) \in \mathbb{R}^{n}$ is the generalized gravity force, $B \in \mathbb{R}^{n \times m}$ maps control inputs into generalized forces, and $J(q) \in \mathbb{R}^{n \times p}$ maps any external forces or constraint forces $F \in \mathbb{R}^{p}$ into generalized forces. 
Common algorithms include: \textit{Forward Dynamics}, computing $\ddot{q}$ when given $q,\dot{q},\tau$, and optionally $F$; \textit{Inverse Dynamics}, computing $\tau$ when given $q,\dot{q},\ddot{q}$ and optionally $F$; as well as the computations of the various terms present in Equation~\ref{eq:manipulator}. %the manipulator equation.

During computation, spatial algebra represents most quantities as operations over vectors in $\mathbb{R}^6$ and matrices in $\mathbb{R}^{6 \times 6}$, defined in the frame of each rigid body. These frames are numbered $i = 1 \text{ to } n$ such that each body's parent $\lambda_i$ is a lower number. Most rigid body dynamics algorithms operate via outward and inward loops over these frames collecting and transforming forces, accelerations, velocities, and inertias. Transformation matrices from frame $\lambda_i$ to $i$ are denoted as ${}^i X_{\lambda_i}$ and can be constructed from the rotation and translation between the two coordinate frames, which themselves are functions of the joint position $q_i$ between those frames and constants derived from the robot's topology. The mass distribution of each link is denoted by its spatial inertia $I_i$, and $S_i$ is a joint-dependent term denoting in which directions a joint can move (and is often a constant). Finally, spatial algebra uses spatial cross product operators $\times$ and $\times^*$, in which a vector is re-ordered into a matrix, and then a standard matrix multiplication is performed. This reordering is shown in Equation~\ref{eq:spatialCross} for a vector $v \in \mathbb{R}^6$:

\setlength{\arraycolsep}{1pt}
\renewcommand\arraystretch{0.5}
\begin{equation} \label{eq:spatialCross}
   \begin{split}
      &v \times = 
      \begin{bmatrix}
            &0 &-v[2] &v[1] &0 &0 &0 &\\
            &v[2] &0 &-v[0] &0 &0 &0 &\\
            &-v[1] &v[0] &0 & 0 &0 &0 &\\
            &0 &-v[5] &v[4] &0 &-v[2] &v[1] \\
            &v[5] &0 &-v[3] &v[2] &0 &-v[0] \\
            &-v[4] &v[3] &0 &-v[1] &v[0] &0 \\
      \end{bmatrix} \\
      &v \times^* = -v \times^T.
   \end{split}
\end{equation}
\renewcommand\arraystretch{1}

For more information on spatial-algebra-based rigid body dynamics we suggest reading Featherstone's \textit{Rigid Body Dynamics Algorithms}~\cite{Featherstone08}.

\section{The GRiD Library} \label{sec:features}
The open-source GRiD library can be found at {\small \href{https://github.com/robot-acceleration/grid}{\color{blue} \texttt{https://github.com/robot-acceleration/grid}}}. In this section we describe its design, features and code optimization approach.

\subsection{Design}
Our overarching design methodology was to make GRiD easily adoptable and extensible by other robotics researchers. As such, the resulting optimized CUDA C++ code is designed to be header-only with only a single dependency, the standard \texttt{cuda\_runtime.h} library.\footnote{Even during URDF parsing and code generation, GRiD only requires the \texttt{beautifulsoup4}, \texttt{lxml}, \texttt{numpy}, and \texttt{sympy} Python libraries.}
%
%The GRiD library is also designed to be used by both GPU experts and novices. As such, we provide an API that allows users to integrate GRiD either directly into their existing CUDA code or through standard CPU C++ function calls. 
%
%We also provide functions to automatically initialize and allocate all necessary memory on the CPU and GPU.
We also provide APIs that allows users to automatically initialize and allocate all necessary memory on the CPU and GPU and integrate GRiD either directly into their existing CUDA code or through standard CPU C++ function calls. 

Finally, the GRiD library is built using a set of modular open-source packages (shown in Figure~\ref{fig:GRiDFig}) to enable easy extension, and re-use by other robotics researchers. Our \textbf{\texttt{GRiD}} package wraps and automates our GPU code generation engine (\textbf{\texttt{GRiDCodeGenerator}}), a self-contained URDF parser (\textbf{\texttt{URDFParser}}), and a set of reference implementations of rigid body dynamics algorithms (\textbf{\texttt{RBDReference}}) that can be used for code validation and testing. 
We also provide the benchmark experiments as described in Section~\ref{sec:benchmarks} as a separate package (\textbf{\texttt{GRiDBenchmarks}}), as they require the support of additional external libraries. 

\iffigs
\begin{figure}[ht]
   \centering
%   \vspace{5pt}
   \includegraphics[width=0.9\columnwidth]{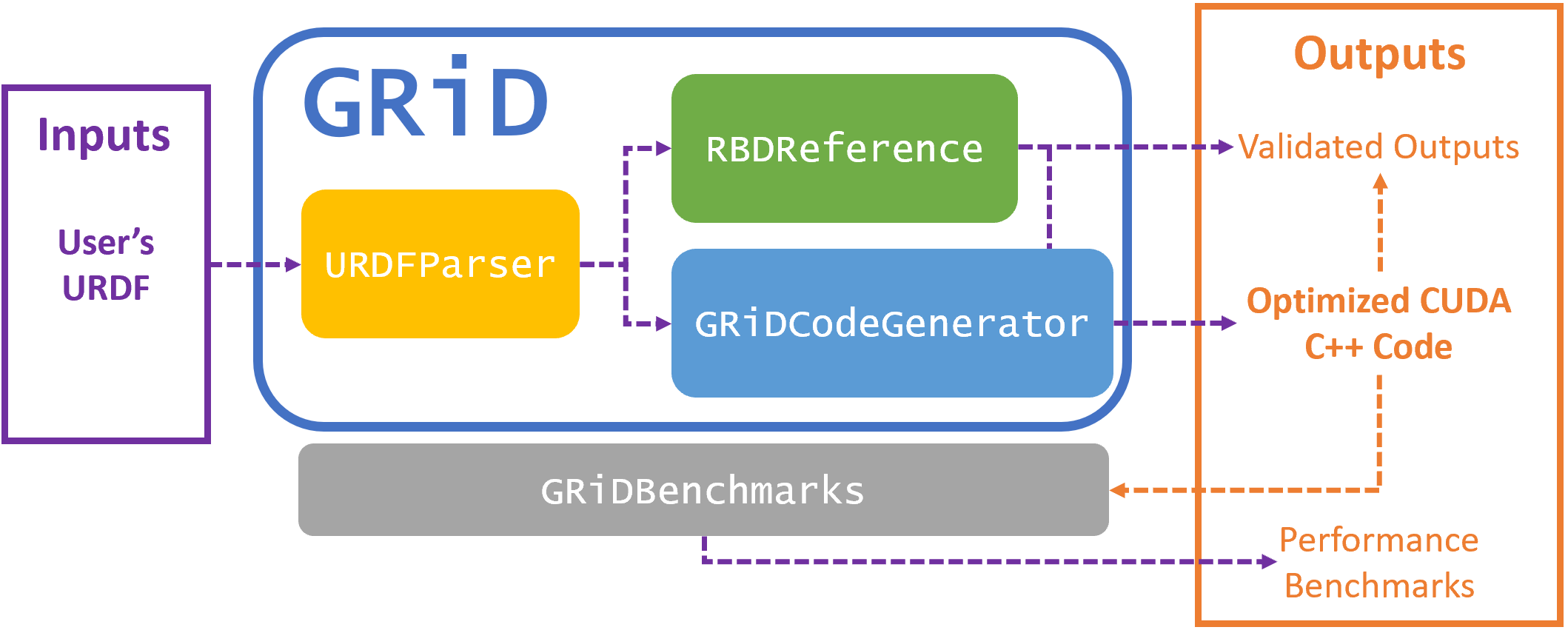}
   \caption{The GRiD library package ecosystem, showing how a user's URDF file can be transformed into optimized CUDA C++ code which can then be validated against reference outputs and benchmarked for performance.}
   \vspace{-5pt}
   \label{fig:GRiDFig}
\end{figure}
\fi
\subsection{Current Features} \label{sec:current_algs}
The GRiD library currently fully supports any robot model consisting of revolute, prismatic, and fixed joints, and implements the following rigid body dynamics algorithms:
\begin{itemize}
    \item The Recursive Newton Euler Algorithm (RNEA) for inverse dynamics~\cite{Featherstone08};
    \item The direct inverse of mass matrix ($M^{-1}$)~\cite{Carpentier18};
    \item Forward dynamics via $-M^{-1}(\tau-\text{RNEA}(q,\dot{q},0))$~\cite{Featherstone08};
    \item The analytical gradient of inverse dynamics with respect to the robot's position and velocity ($q, \dot{q}$)~\cite{Carpentier18};
    \item The analytical gradient of forward dynamics with respect to the robot's position, velocity, and input torque ($q, \dot{q}, u$) via $\frac{\partial \ddot{q}}{\partial u} = -M^{-1} \frac{\partial \text{RNEA}(q,\dot{q},\ddot{q})}{\partial u}$~\cite{Carpentier18}.
\end{itemize}
Directions for future work include extending this core with additional algorithms and joint types (see Section~\ref{sec:conclusion}).
\subsection{Code Optimization Approach} 

GRiD builds on recent work which designed a manually-optimized GPU implementation of rigid body dynamics gradients for a seven-link serial chain manipulator~\cite{Plancher21}.
This section details how GRiD leverages the optimizations identified in that prior work, and how GRiD extends and generalizes those optimizations to enable acceleration of
%In this section we detail the previous optimizations we leverage, as well as the generalizations, and subsequent optimizations, that enable GRiD to accelerate 
both a much larger class of robot models and additional rigid body dynamics algorithms through \texttt{URDF} driven code generation.

Previous work has shown that a robot's topology and joint types directly define structured sparsity patterns and opportunities for parallelism in the resulting spatial-algebra-based rigid body dynamics algorithms~\cite{Featherstone10,Frigerio16,Carpentier19,Plancher21,Neuman21}.
GRiD leverages coarse-grained parallelism using multi-threading between independent computations, as well as \emph{fine-grained parallelism within each computation}.
%We leverage this parallelism using multi-threading not only between independent computations, but at a fine granularity \emph{within} computations.
For example, within each independent gradient computation, each column of that computation can be computed in parallel. Similarly, within those computations each entry in each matrix-matrix or matrix-vector multiplication can be computed in parallel. GRiD is able to achieve high performance by taking advantage of this fine-grained parallelism, through the use of blocks of parallel threads on a GPU.

In order for a GPU to effectively take advantage of such fine-grained parallelism, however, previous work also demonstrated that the target algorithm needs to be refactored to remove synchronization points, and to coalesce both memory accesses and computational operations~\cite{Plancher21}.
This is particularly important for the spatial cross product operations (Equation~\ref{eq:spatialCross})   
which result in out-of-order memory accesses and an expansion of the dimension of the input vector into an output matrix.
We adapt the refactoring approach used in prior work~\cite{Plancher21}, moving computations out of serial loops by creating temporary variables that can be computed in parallel.

\iffigs
\begin{figure}[!b]
   \centering
   \vspace{-15pt}
   \includegraphics[width=0.6\columnwidth]{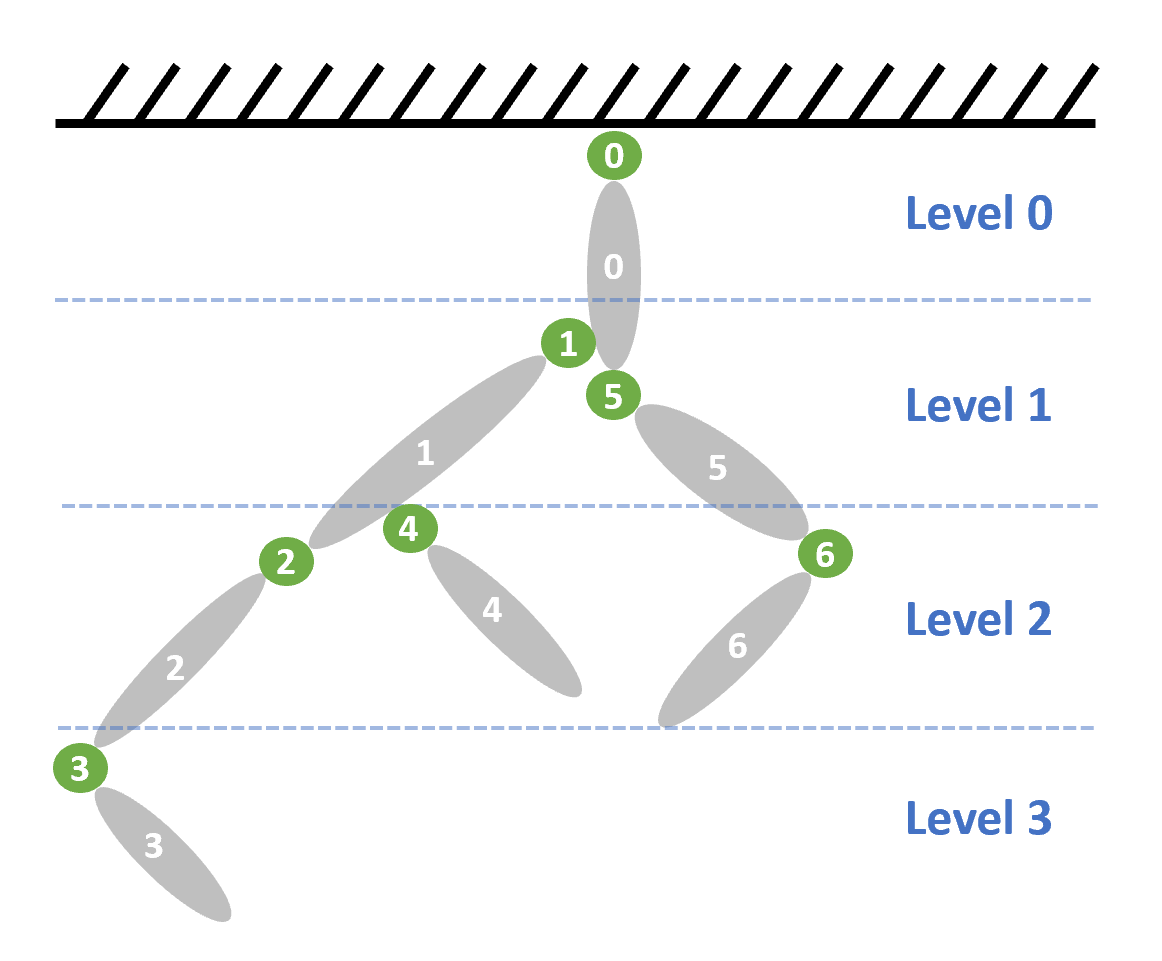}
   \caption{An example robot topology.}
   \label{fig:random_robot}
%   \vspace{-5pt}
\end{figure}
\fi

GRiD extends and generalizes these parallelism-generating optimizations, enabling it to target any robot with a branched tree topology (e.g., Figure~\ref{fig:random_robot}).
To do this, we inject additional optimizations to accommodate multiple branching points at different levels of the tree.
For example, since dependencies in the serial passes of rigid body dynamics algorithms are between parent and child frames in the tree, we can compute ``sibling'' frames in parallel.
For example, the forward pass of $\nabla$RNEA (Algorithm~\ref{alg:RNEAGrad}) computes the temporary variables $\partial v_i, \partial a_i$ for frame~$i$ as a function of $\partial v_{\lambda_i}, \partial a_{\lambda_i}$ for its parent frame~$\lambda_i$  (Lines 2 and 3).
Therefore, we can compute each $\partial v_i, \partial a_i$ by stepping serially through the levels of the tree, while computing all frames within each level in parallel.
For the robot shown in Figure~\ref{fig:random_robot}, we would compute the values associated with frame 0, then 1 and 5 in parallel, then 2, 4, and 6 in parallel, and finally 3.
GRiD also performs loop unrolling on these remaining serial loops to enable the compiler to easily optimize the resulting code.
Then, once all $\partial v, \partial a$ have been computed, all $\partial f$ can be computed fully in parallel.

%%%%%%%%%%%%%%%%%%%%%%%%%%%%%%%%%%%%%%%%%%%%%%%%%%%%%%%%%%%%%%%%%%%%%%%%%%%%%%%%%%%%%%%%%%%%%%%%%%%%%%%%%%%%%%%%%%%%%%%%%%%%%%%%%%%%%%%%

\floatstyle{spaceruled}% Select new float style
\restylefloat{algorithm}% Apply spaceruled float style to algorithm
\begin{algorithm}[!t]
\begin{spacing}{1.25}
\begin{algorithmic}[1]
\caption{$\nabla$RNEA-F($\dot{q},v,a,f,X,S,I$) $\rightarrow$ $\partial c/\partial u$} 
\label{alg:RNEAGrad}
\For{frame $i = 1:N$}
   \State $\frac{\partial v_i}{\partial u} = {}^iX_{\lambda_i} \frac{\partial v_{\lambda_i}}{\partial u} + \begin{cases} \left({}^iX_{\lambda_i} v_{\lambda_i}\right)\times S_i & \phantom{.....} u \equiv q\\ S_i & \phantom{.....} u \equiv \dot{q} \end{cases}$
   \State $\frac{\partial a_i}{\partial u} = {}^iX_{\lambda_i} \frac{\partial a_{\lambda_i}}{\partial u} + \frac{\partial v_{\lambda_i} }{\partial u} \times S_i \dot{q}_i + \begin{cases} \left({}^iX_{\lambda_i} a_{\lambda_i}\right)\times S_i\\ v_i \times S_i \end{cases}$
   \State $\frac{\partial f_i}{\partial u} = I_i \frac{\partial a_i}{\partial u} + \frac{\partial v_i}{\partial u} \times^* I_i v_i + v_i \times^* I_i \frac{\partial v_i}{\partial u}$
\EndFor
% \For{frame $i = N:1$}
%   \State $\frac{\partial c_i}{\partial u} = S_i^T \frac{\partial f_i}{\partial u}$
%   \State $\frac{\partial f_{\lambda_i}}{\partial u} \mathrel{+}= {}^iX_{\lambda_i}^T \frac{\partial f_i}{\partial u} + {}^iX_{\lambda_i}^T \left(S_i \times^* f_i\right)$
% \EndFor
\end{algorithmic}
\end{spacing}
\end{algorithm}

%%%%%%%%%%%%%%%%%%%%%%%%%%%%%%%%%%%%%%%%%%%%%%%%%%%%%%%%%%%%%%%%%%%%%%%%%%%%%%%%%%%%%%%%%%%%%%%%%%%%%%%%%%%%%%%%%%%%%%%%%%%%%%%%%%%%%%%%

\floatstyle{spaceruled}% Select new float style
\restylefloat{algorithm}% Apply spaceruled float style to algorithm
\begin{algorithm}[!t]
\begin{spacing}{1.25}
\begin{algorithmic}[1]
\caption{$\nabla$RNEA-F-GRiD($\dot{q},v,a,f,X,S,I$) $\rightarrow$ $\partial f/\partial u$} \label{alg:RNEAGradGRiD}
\For{frame $i = 1:n$ \textbf{in parallel}}
   %%%%% part 1 - matVMult
    \State $\alpha_i = {}^iX_{\lambda_i} v_{\lambda_i} \hspace{10pt}
            \beta_i = {}^iX_{\lambda_i} a_{\lambda_i}  \hspace{9pt}
            \gamma_i = I_i v_i$
    %%%%% part 2 - mx fx
    \State $\alpha_i = \alpha_i \times S_i \hspace{11.5pt}
           \beta_i = \beta_i \times S_i   \hspace{13pt}
           \delta_i = v_i \times S_i$
\EndFor
\For{level $l = 0:l_{max}$}
    \For{frame $i \in l$ \textbf{in parallel}}
       \State $\frac{\partial v_i}{\partial u} = {}^iX_{\lambda_i} \frac{\partial v_{\lambda_i}}{\partial u} + \begin{cases} \alpha_i \quad \quad u \equiv q \\ S_i \quad \quad u \equiv \dot{q} \end{cases}$
    \EndFor
\EndFor
\For{frame $i = 1:n$ \textbf{in parallel}}
   \State $\rho_i = \frac{\partial v_{\lambda_i} }{\partial u} \times S_i \dot{q}_i + \begin{cases} \beta_i \\ \delta_i \end{cases}$
\EndFor
\For{level $l = 0:l_{max}$}
    \For{frame $i \in l$ \textbf{in parallel}}
        \State $\frac{\partial a_i}{\partial u} = {}^iX_{\lambda_i} \frac{\partial a_{\lambda_i}}{\partial u} + \rho_i$
    \EndFor    
\EndFor
\For{frame $i = 1:n$ \textbf{in parallel}}
   \State $\frac{\partial f_i}{\partial u} = \frac{\partial v_i}{\partial u} \times^* \gamma_i \hspace{15pt} \eta_i = v_i \times^* I_i$
   \State $\frac{\partial f_i}{\partial u} = \frac{\partial f_i}{\partial u} + I_i \frac{\partial a_i}{\partial u} + \eta_i \frac{\partial v_i}{\partial u}$
\EndFor
% \For{level $l = 0:l_{max}$}
%     \For{frame $i \in l$ \textbf{in parallel}}
%         \State $\frac{\partial f_{\lambda_i}}{\partial u} \mathrel{+}= {}^iX_{\lambda_i}^T \frac{\partial f_i}{\partial u} + \zeta_i$
%     \EndFor
% \EndFor
% \For{frame $i = n:1$ \textbf{in parallel}}
%     \State $\frac{\partial c_i}{\partial u} = S_i^T \frac{\partial f_i}{\partial u}$
% \EndFor
\end{algorithmic}
\end{spacing}
\end{algorithm}

When supporting arbitrarily large robots it is also important to ensure that the temporary variables fit into the GPU cache.
At code generation time, GRiD determines if it is necessary to forgo any temporary memory computations in order to support robots with many degrees-of-freedom (dof).
For example, for the 30 dof Atlas humanoid, GRiD does not compute each $v \times$ matrix in parallel and then use threaded matrix multiplication (as in previous work~\cite{Plancher21}), but instead computes $v_1\times v_2$ in a few parallel threads, trading off a slight latency penalty for a large savings in shared memory usage.
This results in the refactored forward pass of the $\nabla$RNEA algorithm shown in Algorithm~\ref{alg:RNEAGradGRiD}.

GRiD also leverages the robot's topology to determine sparsity patterns in the many temporary variables needed for the gradient computations.
As such, columns of temporary memory variables that would be all zeros are skipped and shared memory is compressed to effectively remove those columns. For most robot models this leads to significant savings. For example, reducing shared memory usage for the the quadruped robot HyQ~\cite{semini11} by more than 60\%.
\footnote{
%While this does complicate the memory addressing schema, 
Most required memory offsets are computed and cached at code generation time. %, minimizing the impact on latency. Furthermore, 
GRiD employs non-branching \emph{if/else} constructs 
(e.g., {\small \texttt{result = flag$*$val1 + !flag$*$val2}}) to avoid the branching performance penalty for any other pointer offsets or control flow switches.
}

Finally, GRiD employs further optimizations for certain classes of robot models. For example, for all single chain robots, the parent's frame number is always one less than the child's. For these robots, the code generated by GRiD will remove any indirect references to the parent (or child) frame number and instead simply subtract (or add) one. 

GRiD applies similar patterns of refactorings, memory compressions, and computational optimizations across all of the algorithms described in Section~\ref{sec:current_algs}.

\section{Performance Benchmarks} \label{sec:benchmarks}
We benchmark the GRiD library against the Pinocchio library~\cite{Carpentier19},\footnote{We used the pinocchio3-preview branch %to ensure we were using 
for the latest optimized code.} 
a state-of-the-art CPU-implementation of rigid body dynamics that supports optimized CPU code generation of both rigid body dynamics and its analytical gradients. Source code accompanying this evaluation can be found at {\small \href{https://github.com/robot-acceleration/GRiDBenchmarks}{\color{blue} \texttt{https://github.com/robot-acceleration/\\GRiDBenchmarks}}}.

%%%%%%%%%%%%%%%%%%%%%%%%%%%%%%%%%%%%%%%%%%%%%%%%%%%%%%%%%%%%%%%%%%%%%%%%%%%%
%%%%%%%%%%%%%%%%%%%%%%%%%%%%%%%%%%%%%%%%%%%%%%%%%%%%%%%%%%%%%%%%%%%%%%%%%%%%
%%%%%%%%%%%%%%%%%%%%%%%%%%%%%%%%%%%%%%%%%%%%%%%%%%%%%%%%%%%%%%%%%%%%%%%%%%%%

\subsection{Methodology}\label{sec:methodology}
% All results were collected on 
We used a high-performance workstation with a $3.8$GHz eight-core Intel Core i7-10700K CPU and a $1.44$GHz NVIDIA GeForce RTX 3080 GPU running Ubuntu 20.04 and CUDA 11.4.\footnote{For clean timing measurements on the CPU, we disabled TurboBoost and fixed the clock frequency to the maximum.
Code was compiled with \texttt{Clang 12} and \texttt{g++9.4}, and time was measured with the Linux system call \texttt{clock\_gettime()}, using \texttt{CLOCK\_MONOTONIC} as the source.}
We compare timing results across three robot models: the 7 degrees-of-freedom (dof) Kuka LBR IIWA-14 manipulator~\cite{Kuka20}, the 12 dof HyQ quadruped~\cite{semini11}, and the 30 dof Atlas humanoid~\cite{Atlas21,Kuindersma16}. For single computation and multiple computation latency, we took the average of one million, and one hundred thousand trials, respectively.
%%%%%%%%%%%%%%%%%%%%%%%%%%%%%%%%%%%%%%%%%%%%%%%%%%%%%%%%%%%%%%%%%%%%%%%%%%%%
%%%%%%%%%%%%%%%%%%%%%%%%%%%%%%%%%%%%%%%%%%%%%%%%%%%%%%%%%%%%%%%%%%%%%%%%%%%%
%%%%%%%%%%%%%%%%%%%%%%%%%%%%%%%%%%%%%%%%%%%%%%%%%%%%%%%%%%%%%%%%%%%%%%%%%%%%

\iffigs
\begin{figure*}[!t]
   \centering
   \includegraphics[width=0.93\textwidth]{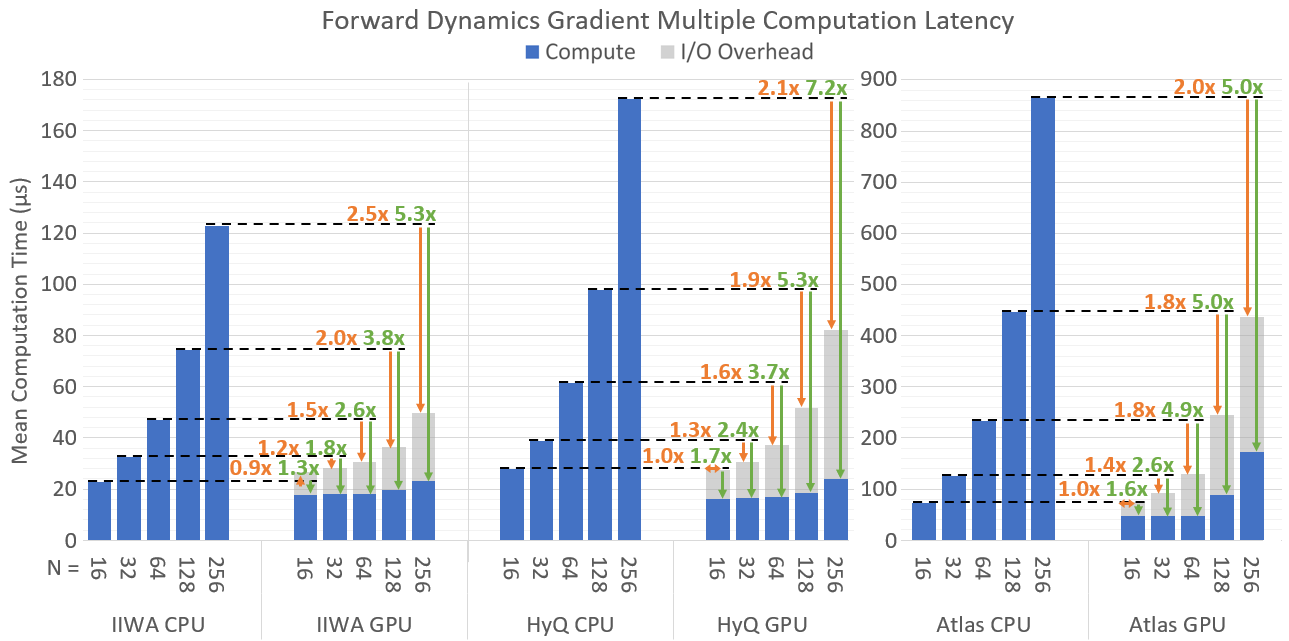}
   \caption{Latency (including GPU I/O overhead) for N = 16, 32, 64, 128, and 256 computations of the gradient of forward dynamics for both the Pinocchio CPU baseline and the GRiD GPU library for various robot models (IIWA, HyQ, and Atlas as described in Section~\ref{sec:methodology}). Overlayed is the speedup (or slowdown) of GRiD as compared to Pinocchio both in terms of pure computation and including I/O overhead.}
   \label{fig:benchmark_multi_fd_grad}
%   \vspace{-5pt}
\end{figure*}
\fi

\subsection{Multiple Computation Latency}
\label{sec:multiple-computation}
To characterize our performance in a typical nonlinear trajectory optimization scenario, which uses tens to hundreds of naturally parallel computations of dynamics algorithms, we evaluate the latency for $N = 16$, $32$, $64$, $128$, and $256$ computations of the gradient of forward dynamics using Pinocchio and GRiD across robot models in Figure~\ref{fig:benchmark_multi_fd_grad}. These times are broken down into computation time on the CPU or GPU and the GPU I/O overhead. The plot is overlayed with the speedup (or slowdown) of GRiD compared to Pinocchio in pure computation alone, and also the speedup including I/O overhead. We use the gradient of forward dynamics as our representative kernel because it uses many of the other kernels as sub-routines and is the most computationally intense kernel, clearly demonstrating scaling trends.

The GPU outperforms the CPU on all but one of the multiple computation latency tests, even when accounting for I/O. In the one test where the CPU is faster---for the fewest computations, including I/O, for IIWA, the smallest robot with only a single limb---the GPU is still 0.9x as fast.

Even on the CPU, this benchmark shows how important it is to take advantage of coarse-grained parallelism between computations. For example, the gradient of forward dynamics kernel ($\nabla FD$ in Table~\ref{tab:single_timing}) took 2.9, 4.3, and 20.9 $\mu$s for a single computation for IIWA, HyQ, and Atlas respectively. If we ran it 256 times serially it would therefore take over 742, 1091, and 5355 $\mu$s. As Figure~\ref{fig:benchmark_multi_fd_grad} shows, computing it in parallel on 8-cores only takes 123, 172, 865 $\mu$s, saving 83-84\% of the computation time. 

However, since the CPU only has 8 cores, it is unable to efficiently scale to take advantage of high numbers of naturally parallel computations, taking 5.4x, 6.2x, and 11.8x as long to compute N = 256 as compared to N = 16 for IIWA, HyQ, and Atlas respectively.

The GPU, on the other hand, is designed to scale to higher numbers of computations without incurring a latency penalty by launching independent blocks of threads for each computation. In fact, for IIWA and HyQ, N = 256 takes only 1.3x and 1.5x as long as N = 16. This leads to the GPU outperforming the CPU by 5.3x and 7.2x for N = 256, and maintaining a 2.5x and 2.1x speedup when including I/O.

For the much higher-dof Atlas robot, the GPU still outperforms the CPU for N = 256 by 5.0x, and 2.0x when including I/O. However, unlike with IIWA and HyQ, this performance increase is almost identical to the increase at N = 128 and N = 64. This stall in performance improvement is caused by the large amount of shared memory needed for Atlas's 30 dof which starts to limit the number of parallel blocks of threads that can fit concurrently on the GPU hardware.

% For both the CPU and GPU, the variance in timing results decreases as the computational complexity increases. The GPU is very reliable with more than 90\% of times within 1\% of the mean for all experiments and achieves over 99\% for $N \geq 64$ for Atlas. The CPU, on the other hand, only has 95\%, 87\%, and 86\% of times within 1\% of the mean for $N = 256$ for Atlas, HyQ, and IIWA respectively, which drops to 81\%, 73\%, and 54\% for $N = 16$.

Finally, we note that for the GPU, I/O overhead accounts for 53-71\% of the total time for N = 256. This indicates that GRiD can provide the highest performance if integrated directly into an entirely GPU-based algorithm, instead of being used to accelerate a step of a CPU-based algorithm. In either case, however, if there is sufficient parallel work to be done, GRiD can reduce the overall computational latency.

%%%%%%%%%%%%%%%%%%%%%%%%%%%%%%%%%%%%%%%%%%%%%%%%%%%%%%%%%%%%%%%%%%%%%%%%%%%%
%%%%%%%%%%%%%%%%%%%%%%%%%%%%%%%%%%%%%%%%%%%%%%%%%%%%%%%%%%%%%%%%%%%%%%%%%%%%
%%%%%%%%%%%%%%%%%%%%%%%%%%%%%%%%%%%%%%%%%%%%%%%%%%%%%%%%%%%%%%%%%%%%%%%%%%%%

\iffigs
\begin{figure*}[!t]
   \centering
   \includegraphics[width=0.9\textwidth]{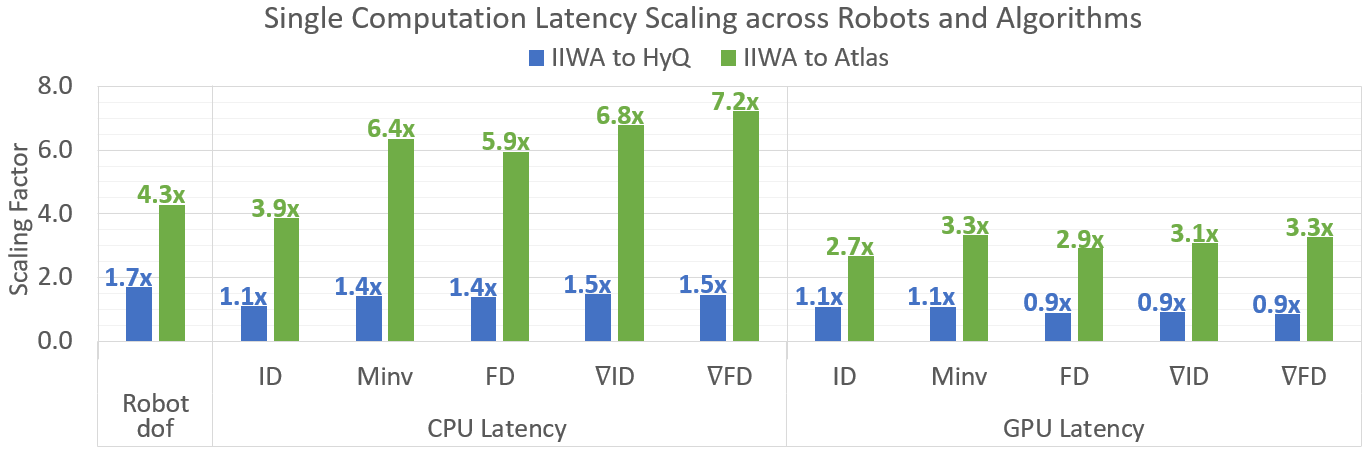}
   \caption{The scaling of single computation latency from IIWA to HyQ and IIWA to Atlas for both the Pinocchio CPU baseline and the GRiD GPU library for various rigid body dynamics algorithms (ID = Inverse Dynamics, Minv = Direct Minv, FD = Forward Dynamics and $\nabla$ indicates the gradient of that algorithm). We also plot the scaling of the robots' dof as a measure of their increased complexity.}
   \label{fig:benchmark_single}
%   \vspace{-5pt}
\end{figure*}
\fi

\begin{table}[!t]
   \footnotesize
   \renewcommand{\arraystretch}{1.3}
   \caption{Single Computation Latency in $\mu$s Per Algorithm and Robot {\scriptsize (ID = Inverse Dynamics, Minv = Direct Minv, FD = Forward Dynamics and $\nabla$ indicates the gradient of that algorithm)}}
   \label{tab:single_timing}
   \centering
   \begin{tabular}{|l|l|l|l|l|l|l|}
        \cline{2-7}
        \nocell{1} & \multicolumn{3}{c|}{CPU} & \multicolumn{3}{c|}{GPU} \\
        \hline
        Algorithm   & IIWA & HyQ & Atlas 
                    & IIWA & HyQ & Atlas
                    \\ \hline\hline
        ID          &  $0.3$ &  $0.3$ & $1.1$ 
                    &  $3.0$ &  $3.2$ & $8.0$
                    \\ \hline
        Minv        &  $0.5$ &  $0.8$ & $3.4$             
                    &  $5.2$ &  $5.6$ & $17.4$
                    \\ \hline
        FD          &  $0.9$ &  $1.2$ & $5.3$
                    &  $7.7$ &  $6.9$ & $22.4$
                    \\ \hline
        $\nabla$ID  &  $1.4$ &  $2.1$ & $9.8$
                    &  $6.3$ &  $5.8$ & $19.5$
                    \\ \hline
        $\nabla$FD  & $2.9$ & $4.3$ & $20.9$
                    & $12.9$ & $11.0$ & $42.1$
                    \\ \hline
   \end{tabular}
   \vspace{-10pt}
\end{table}

\subsection{Single Computation Latency Scaling}
\label{sec:single-computation}

For further analysis of the GPU performance benefits demonstrated in Section~\ref{sec:multiple-computation}, we plot the latency scaling, excluding I/O overheads, of a single computation of each rigid body dynamics algorithm, from IIWA to HyQ and IIWA to Atlas, on both the CPU and GPU in Figure~\ref{fig:benchmark_single}, and list absolute timings in Table~\ref{tab:single_timing}. We also plot the scaling of the robots' dof as a measure of their computational complexity.

We find that the GPU is able to scale to more complex robots and algorithms better than the CPU by taking advantage of fine-grained parallelism induced by independent robot limbs and the independent columns of gradient computations. 
%This is why GPUs are most advantageous when used on computations where they can take advantage of both fine and coarse-grained parallelism. 
As expected (and consistent with previous work~\cite{Plancher21}), the GPU is slower on a single computation than the CPU, but the GPU demonstrates better scalability across both algorithm and robot complexity. 
For example, as shown in Table~\ref{tab:single_timing}, for $\nabla FD$, the CPU is 4.4x faster than the GPU (2.9$\mu$s vs. 12.9$\mu$s), but only 2.0x faster for Atlas (20.9$\mu$s vs. 42.1$\mu$s). 
% For example, for $\nabla FD$, the GPU and CPU take 12.9$\mu$s versus 2.9$\mu$s for IIWA--a difference of 4.4x--but for the more complex Atlas, they take only 42.1$\mu$s versus 20.9$\mu$s--a significantly reduced difference of 2.0x (Table~\ref{tab:single_timing}).
That said, the CPU is still faster than the GPU for all individual computations, showing that GPU acceleration only makes sense when there is sufficient parallel work to be done.

On the CPU, the latency of each algorithm scales directly with its computational intensity, with the gradients requiring significantly more computation (see Table~\ref{tab:single_timing}). 
The most computationally intensive algorithm, the forward dynamics gradient ($\nabla FD$), takes 2.9, 4.3, and 20.9 $\mu$s for IIWA, HyQ, and Atlas, while the simplest algorithm, inverse dynamics ($ID$) takes 0.3, 0.3, and 1.1 $\mu$s---a 9.7x to 19.0x slowdown.

CPU latency also scales with the dof of the robot (Figure~\ref{fig:benchmark_single}). For example, as the robot's dof increases by a factor of 1.7x from IIWA to HyQ, the computation time also increase by 1.1x, for the $O(N)$ $ID$ algorithm, up to 1.5x, for the $O(N^2)$ $\nabla FD$ algorithm. It appears that this strong performance is due to the code generation taking advantage of the the many shared computations in the gradients, as well as the sparsity induced by HyQ's independent limbs, which decrease the longest path through the rigid body tree from 7 on IIWA to 3 on HyQ.
However, these optimizations are mitigated by the Atlas model, which has a much larger 30 dof, and a longest path through the rigid body tree of 8.
Atlas has 4.3x the dof of IIWA, but has a 3.9x ($ID$) to 7.2x ($\nabla FD$) slowdown on the CPU.

By contrast, the GPU is able improve its scalability by not only taking advantage of sparsity and shared computations, but \emph{also the opportunities for fine-grained parallelism} caused by both independent limbs in complex robot models, and independent columns of the gradient computations.
For example, Table~\ref{tab:single_timing} shows that by taking advantage of parallelism in the gradient computations,
the GPU is not only able to compute $\nabla ID$ faster than $FD$, but also 
only takes 12.9, 11.0, and 42.1 $\mu$s (for IIWA, HyQ, Atlas) for $\nabla FD$ as compared to 3.0, 3.2, and 8.0 $\mu$s for $ID$---a slowdown of only 3.4x to 5.3x, and a significant reduction from the CPU's 9.7x to 19.0x slowdown for these algorithms.
Similarly, Figure~\ref{fig:benchmark_single} shows that by leveraging limb-based parallelism, the GPU computes forward dynamics ($FD$) and both gradients ($\nabla ID, \nabla FD$) faster for HyQ than for IIWA, and only has a 2.7x to 3.3x slowdown from IIWA to Atlas, again a significant reduction from the CPU's 3.9x to 7.2x.

\section{Conclusion and Future Work} \label{sec:conclusion}
In this work, we introduce GRiD, a GPU-accelerated rigid body dynamics library with analytical gradients.
We found that by leveraging large-scale parallelism when performing multiple computations of rigid body dynamics algorithms, GRiD can provide as much as a 7.2x speedup over a state-of-the-art, multi-threaded CPU implementation and maintains as much as a 2.5x speedup when including I/O overhead.

There are many promising directions for future work to extend the functionality and versatility of the GRiD library. 
We have current work under development to expand GRiD to support the full breadth of rigid body dynamics algorithms and robot models supported by current state-of-the-art CPU spatial-algebra-based rigid body dynamics libraries~\cite{Todorov12,Frigerio16,Carpentier19,Koolen19,Werling2021}.
Additionally, we are developing wrappers to our C++ host functions in higher level languages to make it even easier to leverage GRiD.

We would like to explore emerging rigid body dynamics algorithms and alternate formulations and implementations of rigid body dynamics, which may improve overall performance by exposing additional parallelism and computational efficiency~\cite{Featherstone99,Yamane09,Yang17,Brudigam20,Nganga21,Singh21,Echeandia21,Agboh19,Agboh20}.

We would also like to add support for differentiating through model parameters~\cite{Heiden2021,Werling21}, as well as for contact, and hope to integrate these accelerated dynamics implementations into existing robotics software frameworks~\cite{Tedrake16,Brockman16,Giftthaler18,Plancher18,Howell19}. This would increase both GRiD's ease-of-use and applicability to more robotics researchers. 

Finally, building out increased support for more trajectory optimization, MPC, and ML algorithms running entirely on the GPU would further increase the performance benefits from integrating GRiD into these approaches.

% \addtolength{\textheight}{-7.5cm}
\bibliographystyle{bib/IEEEtran}
\bibliography{bib/IEEEabrv,bib/agile.bib}

\end{document}

\end{document}